# Resilient conductive membrane synthesized by in-situ polymerisation for wearable non-invasive electronics on moving appendages of cyborg insect

Qifeng Lin[1#], Rui Li[1#], Feilong Zhang[2#], Kai Kazuki[1], Ong Zong Chen[1], Xiaodong Chen[2,*], Hirotaka Sato[1,*]

[1]School of Mechanical & Aerospace Engineering, Nanyang Technological University; 50 Nanyang Avenue, 639798, Singapore

[2]Innovative Center for Flexible Devices (iFLEX), Max Planck–NTU Joint Lab for Artificial Senses, School of Materials Science and Engineering, Nanyang Technological University, 50 Nanyang Avenue, Singapore 639798, Singapore

[#]These authors contribute equally to this work.

[*]Corresponding author. Email: chenxd@ntu.edu.sg; hirosato@ntu.edu.sg

**Abstract**

By leveraging their high mobility and small size, insects have been combined with microcontrollers to build up cyborg insects for various practical applications. Unfortunately, all current cyborg insects rely on implanted electrodes to control their movement, which causes irreversible damage to their organs and muscles. Here, we develop a non-invasive method for cyborg insects to address above issues, using a conformal electrode with an in-situ polymerized ion-conducting layer and an electron-conducting layer. The neural and locomotion responses to the electrical inductions verify the efficient communication between insects and controllers by the non-invasive method. The precise "S" line following of the cyborg insect further demonstrates its potential in practical navigation. The conformal non-invasive electrodes keep the intactness of the insects used while controlling their motion. With the antennae, important olfactory organs of insects preserved, the cyborg insect, in the future, may be endowed with abilities to detect the surrounding environment.

**Introduction**

Centimeter-scale mobile robots recently received increasing attention due to the growing demand to work in confined spaces, such as exploration in urban post-disaster terrain and pipeline inspection[1–4]. Current robots at this scale mainly include robots composed of mechanical structures[1,4–6] and cyborg insects[2,7–9]. Despite the advantages of dynamics control, the former generally suffer from high energy consumption and poor adaptability to complex environments. In contrast, cyborg insects preserve the insect's locomotion ability and adaptability to the environment, thus exhibiting low energy requirements[10]. As such, cyborg insects have attracted tremendous attention and are promising for practical applications. Current research about cyborg insects focuses more on remote and precise control over insects' locomotion [7,11,12]. Many factors could induce insects' motion, such as light, heat, chemicals, and so on[13–16]. However, the prementioned factors have not been considered in the research, as the stimulation protocol should remain fully onboard and not inhibit the insect's wider motion range. Moreover, due to the ease of mounting the electronic stimulus-releasing module and adjusting the intensity of the stimulus, electrical stimulation via wire implanted in the insects' body was the only verified stimulation protocol for the research on the cyborg insect[10–12,17–19].

However, this invasive electrode implantation severely damages the functional organs, such as antennae and cerci, and disrupts the insect's physiology (Fig. 1a)[7,11]. As the antennae are a vital olfactory organ and directly influence the contact between insects and surrounding objects, damage to the antennae will affect the insect's ability to perceive its surroundings and its natural self-avoidance behaviour against obstacles[20,21]. The electrodes implanted inside the insect's body may cause damage to its internal organs and potentially shorten its lifespan. Additionally, the long-term stable control of the cyborg insect could also be undermined.

Thus, it is urgent to develop non-invasive electrodes to guide insect movements without compromising their physical intactness or natural ability to interact with the environment.

Madagascar Hissing Cockroach (Fig. 1b) has been extensively used for cyborg insects by virtue its high load capacity (15g), however, control with non-invasive electrodes still face two main challenges. One is to create a robust interface to communicate with the insect's moving appendages. The robust interface requires the electrodes to be adhered to the insect body surface to guarantee low interfacial impedance. At the same time, the electrodes should not prevent the regular movement and sensitivity of the insect's specific appendages. For example, the cockroach's antenna is full of hairs and freely moves to probe the environment, which make it difficult to realize close and secure contact with the electrodes difficult. The other challenge is the long-term stability of the electrodes and interfaces. Considering the possible harsh environment and the need for extended working time, the electrodes must be electrically and mechanically stable under a complex and volatile environment to ensure adequate contact with the insect body surface.

Here, we report a novel cyborg insect with non-invasive wearable electrodes realizing the precise control of the insect's locomotion. The non-invasive electrodes comprise an ion-conducting layer of poly ionic liquid (PIL) gels interfacing with the insect tissue surface and a electron-conducting layer of gold on PI film or thin silver wire connecting the controller. Thus, the electrical stimulation can be transferred to the antennae and abdominal part of the insect (cockroach). The PIL gels formed by in-situ photopolymerisation after the precursors wet the surfaces realizing conformal and robust interfaces with the antenna and abdominal skin of the cockroach for stable electrical communications. With the non-invasive electrodes, this cyborg insect preserves the intactness of the insect's physiological structure and thus maximises the insect's natural movement and perception under the premise of effectively controlling the insect's moving ability.

**Results and Discussion**

To control cyborg insects without harming their tissues, non-invasive electrodes are needed to stimulate the insects through their body surface. The antennae can sense the surrounding obstacles and induce the turning reactions of insect to avoid them [7,11,21,22]. Hence, the antennae are selected for non-invasive electrodes to attached on for triggering the insect's turning reaction (Fig. 1c,d,f). Besides, the tactile stimulation on the insect's dorsal abdomen effectively induced the acceleration reaction [23]. Electrical stimulation to the dorsal abdomen can be utilized to induce the insect's acceleration reaction by placing the non-invasive electrodes inside the inter-spaces of the insect's abdominal segments (Fig. 1g). Compared with the conventional electrode implantation method, the newly designed wearable non-invasive electrodes would not damage the insect's body part (Fig. 1e,g).

Compared with the inter-space surface of the insect's abdominal segments, the antennae had a more flexible range of motion and a hairy and hydrophobic skin surface(Supplementary Fig. 1a,b), which hindered the formation of an intimate interface with conventional hydrogel-based electrodes. Simultaneously, the insect's antenna and abdomen with electrodes must be able to move freely to preserve automatic functions. Therefore, conformal electrodes for cyborg insects should be non-invasive, compliant with the insect surface and movement, and have long-term effects at the electron-ion transduction for electrical stimulation.

The non-invasive conformal electrodes we develop for turning stimulation on the antennae surface comprise an ionic layer of PIL gel connecting the antennae and gold (Au) nanofilm deposited on polyimide (PI) film for electronic transduction. Considering the hairy structures and hydrophobic surfaces of the antennae, hydrophobic PIL precursors with (2-Acryloyloxyethyl)trimethylammonium bis(trifluoromethane)sulfonimide ([AETMA][TFSI]) as the monomer, butyltrimethylammonium bis(trifluoromethanesulfonyl) imide ([N4111][TFSI]) as the solvent and ion carrier, and 2-hydroxy-2-methyl-1-phenylacetone as the photo-initiator were used for the fabrication of PIL gels, which can seep into the hairs and wet the antenna surface (Supplementary Fig. 1c)[24]. The content of [N4111][TFSI] in the gels shows a significant effect on the mechanical and electrical properties of the PIL gels (Fig. 2a, 2b). Higher content of [N4111][TFSI] makes the PIL gels softer and more stretchable with the decreased glass transition temperature but also decreases the strength of the PIL gels (Fig. 2a and Supplementary Fig. 2b). On the other hand, the conductivity of the PIL gels increased when the solvent content increased and reached a plateau at the ratio of the monomer to the solvent ($n_{monomer} : n_{solvent}$) of 6 : 2 (Fig. 2b, impedance at 40 Hz, and Supplementary Fig. 2c). Taking mechanical and electrical properties of PIL gels into consideration, we selected PIL gel with $n_{monomer} : n_{solvent}$ of 6 : 2 for subsequent experiments.

Due to the hair structures of the antenna, it is difficult for the Au-PI film and the prepared solid PIL gel film to realize close contact with the surface of the antenna (Fig. 2c). Thus, we propose to in situ photopolymerize the PIL gel precursors after wetting the antenna and Au-PI film to achieve close contact with the antenna (Supplementary Fig. 1c). As shown in Fig. 2c, the in-situ PIL gel had complete contact with the antenna surface, while there are lots of gaps between the solid gel film and antenna. Therefore, in-situ polymerization method was used for subsequent electrodes. The in-situ polymerization was initiated with UV and the UV exposure duration required is characterized by changes in adhesive force between the PIL gel and antennae (Fig. 2d). The adhesion force plateaued after around 100 s of UV exposure, which therefore is chosen as the exposure time for subsequent electrode preparation. The tangential adhesion force of the antenna-PIL gel-(Au-PI) film system can reach ~100 kPa, which ensured the mechanical stability of the interface (Fig. 2e). This PIL gel is quite stable, which can withstand temperatures up to 400 ℃ (Supplementary Fig. 3a). After 140 days in air, the impedance of the PIL gel remains essentially unchanged (Fig. 2f). Additionally, this PIL gel also remains stable under different levels of humidity (0-100%) and even underwater, with a relative change in impedance of less than 60% (Supplementary Fig. 3b, c). The reliable electrical properties of this PIL gel ensure long-term stability of the non-invasive electrodes for their operation.

To trigger the insect's acceleration reaction, silver wires were placed inside the inter-space between the two neighbouring segments of the insect's abdomen. However, as the insect's abdomen could contract if only silver wire is used, it is difficult to ensure that it remains in contact with the insect's skin surface. Thus, PIL gel was also applied to connect the insect body surface with silver wire at the inter-space between abdominal segments using in-situ polymerisation. Moreover, the silver wire was fixed on the segment with tape to ensure it could not be pulled away.

In order to verify the feasibility of non-invasive electrodes for cyborg insects, neural activities in the neck neuron were recorded under the electrical stimulation with non-invasive electrodes on the antennae (Fig. 3, Supplementary Fig. 4). Different volts (from 1 V to 5 V) were used to stimulate the insects through the non-invasive electrodes and their neural reaction were monitored simultaneously[7]. Neural reaction spikes were detected after electrical stimulation (Fig. 3b). The number of spikes increases with the increase of voltage and plateu at the voltage of 4 V (Fig. 3c) and the neural reaction spikes tend to be stable after the stimulus voltage increases over 2.5 V (Fig. 3d). As control, electrical stimulation with Au-PI film only elicit a few neural reaction spikes and the reactions were weak and unstable (Fig. 3b-d). These results confirm the feasibility of non-invasive stimulation for cyborg insect and the necessity of PIL gel for the non-invasive electrodes. Based on the above results, 4 V electrical stimulus with PIL gel applied on the antennae was used for inducing the turning reaction of the cyborg insect.

After testing the insect's locomotion, the stimulation with an amplitude of 4 V, frequency of 42 Hz and stimulation duration of 1 s could obviously elicit the cyborg insect's reaction for turning and acceleration. Thus, the cyborg insect's actual locomotion control was tested with these parameters. Ten insects were tested to confirm the control performance. Each insect was tested with one trial. Based on the designed stimulation sequence and the Vicon recorded data, the stimulation effect of the non-invasive electrode on the insect could be identified (Fig. 4a). In order to quantify the stimulation effect, criteria were set for turning and acceleration stimulation. If the insect's motion induced by the stimulation met the criteria, then the stimulation of this time would be labelled as success. Otherwise, it would be labelled as a failure. Criteria for success stimulation were set up as follows. For left-turning stimulation, the average angular velocity should be counterclockwise. Conversely, for right-turning stimulation, it should be clockwise. For both cases, if the insects are already turning in the correct direction before stimulation, their angular velocity should be 1.2 times higher after stimulation. For acceleration stimulation, the average linear velocity during the stimulation should be 1.2 times higher than before stimulation.

To determine the performance of locomotion control, 10 cyborg insects were prepared to test and for each of them, 15 forward /left /right stimulation were tested and studied. Thus, 150 forward /left /right stimulation in total were collected. Based on the criteria above, the success rates of the three types of stimulation were identified. Both left-turning and right-turning stimulation achieved an 80.7% success rate, and acceleration stimulation achieved a 74.0% success rate. Meanwhile, the mean value of induced linear and angular velocity was identified. The induced linear velocity increased by nearly 60 mm/s during the acceleration stimulation (Fig. 4b). The angular velocity from the turning stimulation could arrive at around 60 degree/s average (Fig. 4d, 4f). Furthermore, after the stimulation duration, all of the induced velocities decreased. The induced turning angle varied from 0 to around 106 degrees (left-turning) or 109 degrees (right-turning) during 1 second of stimulation (Fig. 4c, 4e). The variation of the turning reaction may be caused by the insect's initial status or individual variability, which was also shown for the other cyborg insects [7,11]. The average induced turning angle was 38.5 degrees for left-turning stimulation and 38.4 degrees for right-turning stimulation. These two average values were very close, with a difference of 0.26%, which showed that the turning stimulation performed similarly to both antennae.

The turning and acceleration stimulation with non-invasive electrodes could effectively induce the insect's motion. Based on the stimulation types and intensity verified before, the cyborg insect was successfully navigated along the path with the shape of "S" manually (Fig. 5). The walking path of the insect was very close to the "S" shape designed, which meant that the non-invasive electrodes implanted insects were easily controlled.

**Conclusions**

We have developed a novel cyborg insect with non-invasive wearable electrodes made of hydrophobic PIL gels by in-situ polymerization. This insect-friendly design preserved the insect's first-ever intactness during the development of the cyborg insects. While keeping the essential sensory organs and antennae undamaged, the locomotion controllability of the cyborg insect was also verified and demonstrated. Nevertheless, the advantages of preserving the insect's body integrity still need to be considered and utilised, especially for the cyborg insect's obstacle negotiation ability. Furthermore, other applications of the novel cyborg insect could be further studied, such as the navigation in complex terrain and the robot's climbing ability on vertical planes.

**Materials and Methods**

*Insect Platform and Non-invasive Electrodes Implant Body Positions*

Among the developed terrestrial cyborg insects, the Male Madagascar hissing cockroach platform had advantages over the suitable load capacity and highly stable and smooth mobility. Besides, the invasive electrodes' acceleration and turning stimulus were studied. Thus, the platform of our cyborg insect was chosen as Male Madagascar hissing cockroach. The insects (5~7.5cm, 6~8g) were fed with carrots and water weekly at the rearing system from the Brand NexGen Mouse 500, Allen Town, with 25 degrees Celsius and 60% of relative humidity. The National Advisory Committee approved the experiment on the insect for Laboratory Animal Research.

In order to steer the insects, antennae were inserted with the metal electrodes for different insect platforms[2,7,11], which showed its effectiveness in inducing the insect's turning motion. Thus, antennae were chosen as the non-invasive electrodes for the turning reaction. Furthermore, as the tip of the antennae was necessary for detecting the surroundings and the moment generated by the electrode's weight at the tip could affect the range of motion of the antenna, the non-invasive electrode was considered to attach to the lower part of the antenna.

As discussed in paper [3,7], the insect's abdominal segment and cerci could be inserted with a grounding electrode and acceleration electrode to induce the insect's acceleration motion. However, the cuticle of the insect abdomen is very smooth to attach the non-invasive electrode. Thus, the inter-space between the abdominal segment was considered for the acceleration and grounding electrodes.

*Non-invasive Electrodes*

The non-invasive electrodes had two parts: poly ionic liquid gel and electron conductor. The electron conductor was directly connected to the electrical stimulus output channel. At the same time, the poly ionic liquid gel supported complete contact between the electron conductor and the insect body surface and conducted the electrical stimulus from the electron conductor. In this paper, the electron conductor could either be a silver wire or a conformable soft electrode, dependent on the choice of the body surface (antenna or inter-space between abdominal segments). This section introduces the preparation of poly ionic liquid gel and electron conductor of the conformable soft electrode. The materials included [2-(Acryloyloxy)ethyl] trimethylammonium chloride solution (AETMAC-Q, 80 wt. % in water, Sigma-Aldrich 496146), bis(trifluoromethane)sulfonimide lithium salt (LiTFSI, Sigma-Aldrich 449504), butyltrimethylammonium bis(trifluoromethylsulfonyl)imide (N1114 TFSI, Sigma-Aldrich 713007), 2-hydroxy-2-methyl-1-phenylacetone (Irgacure1173, Sigma-Aldrich 405655) and acetone, which were all used as received.

*Ionic Liquid Monomer*

The monomer was s synthesized as previously reported[26]. Briefy, 10 g AETMAC-Q was mixed with 40 mL LiTFSI aqueous solution (1 mol L-1) under vigorous stirring for 2 hours. The ionic liquid monomer, [2-(Acryloyloxy)ethyl] trimethylammonium bis(trifluoromethane)sulfonimide ([AETMA][TFSI]) formed as a heavy oil layer after the phase separation facilitated by centrifugation. The oil layer was collected and washed with water five times. Then, the [AETMA] [TFSI] was obtained after vacuum drying at 70 ℃ for 12 hours.

*Poly Ionic Liquid Gel and Its Property Testing*

The poly ionic liquid (PIL) gel films with a specific monomer ratio were synthesised by free-radical polymerisation in the spacer (0.5 mm) of two separated glass slides with [AETMA][TFSI] as the monomer, [N1114][TFSI] as the solvent, and 2-hydroxy-2-methyl-1-phenylacetone as the photo-initiator. Different monomer ratios were tested to study the material's mechanical and electrical properties and to prepare the most suitable one for the insect. Typically, 6 mmol [AETMA] [TFSI], 2 mmol (or 0, 1, 3, 4 mmol) [N1114][TFSI], and 10 μL 2-hydroxy-2-methyl-1-phenylacetone were mixed and bubbled with nitrogen to remove the dissolved oxygen. Then, the polymerisation process proceeded under UV light (365 nm, 80 W) for 20 min to obtain the PIL gel film.

The PIL gel film elongation test was conducted using a commercial testing machine (C42, MTS Systems Corporation). First, the PIL gel films were cut into belts with widths of 5 mm. Then, the samples were clamped at both ends and elongated at a loading rate of 50% s$^{-1}$.

The conductivity of the PIL gels with different monomer ratios was measured by an electrochemical workstation (Zennium E, Zahner Ennium). The samples were prepared as a sandwich structure with the gel film (4 mm × 4 mm) between two electron-conductive layers (Au-PI). The impedance was measured from 1 to $10^6$ Hz. In order to investigate the electrical stability of the PIL gels, the samples were kept in a confined box with different ambient humidity (balanced with different saturated aqueous solutions) for more than two days to reach equilibrium or in water, and then the impedance was recorded.

The PIL gel was fabricated with an optimal suitable molar ratio of [AETMA] [TFSI] to [N1114][TFSI] in two steps. First, 6 mmol [AETMA] [TFSI], certain mmol [N1114][TFSI], and 10 μL 2-hydroxy-2-methyl-1-phenylacetone were dissolved in acetone to 6 mL and polymerised under UV light (365 nm, 80 W) for 20 min. Then, a viscous solution was obtained. After that, another 6 mmol [AETMA] [TFSI], certain mmol [N1114][TFSI], and 10 μL 2-hydroxy-2-methyl-1-phenylacetone were thoroughly mixed in the above viscous solution. The PIL gel was obtained after vacuum drying the solution at 40 ℃ for 12 hours to remove acetone. The specific volume of [N1114][TFSI] was decided during the comparison or the identification process for the best suitable molar ratio of [AETMA] [TFSI] to [N1114][TFSI].

*Fabrication of Conformable Soft Electrode*

The conformable soft electrodes were fabricated by physical vapour deposition of gold (Au) on polyimide (PI) films. Before Au depositions, the PI films were treated with oxygen plasma to enhance the interfacial strength of the Au and PI films. Then, the Au was deposited using a thermal evaporation machine (Nano 36, Kurt J. Lesker) at the rate of 0.4 Å s-1 with the final Au thickness of gold of 70 nm, defined as Au-PI.

*Recording of Neural Reaction against Electrical Stimulation*

As the insect's abdominal segments could wag slightly relative to their next anterior neighbors during the insect's moving [23], the electrodes inside the inter-space must have the PIL gel to keep contact with the insect's skin surface, otherwise, the silver wires quickly lose contact with the skin surface. On the other hand, engineering methods, such as tightening with a fastener, could be used for the antennae to secure the contact between the electrode and skin surface. Thus, to confirm the necessity of PIL gel on the antennae, the insect's neural reactions against the electrical stimulation without and with PIL gel were recorded.

For neural recording, a cockroach was anesthetised with $CO_2$ before dissection and fixed to a wooden platform with ventral side up using insect pins (Fig. 3a). Both Antennae were fixed to a U-shaped pin placed above the scape using UV-glue so that the antennae were immobilised during the experiment. A small incision was made in the neck's ventral cuticle, and internal tissues were removed to expose the ventral nerve cords (VNC). The VNCs were rinsed with cockroach saline for visualisation (139 mM/L NaCl, 5 mM/L KCl, 4mM/L $CaCl_2$[25]).

Glass suction electrodes were used for neural recording. The tip of the glass pipettes was cut and blunt by flaming to get a better fit to a nerve cord. The shaft of the pipette was filled with cockroach saline. Sharp scissors were used to cut the right VNC, and the proximal end was sucked into the glass electrode. Ag-AgCl wire was used for recording and reference electrodes, and the reference electrode was put near the tip of the suction electrode. A ground pin was inserted into the insect's abdomen. Recordings were made with an extracellular amplifier through a 300 Hz high-pass and 20 kHz low-pass filter (Model 3000, AM-Systems,

USA). Signals were digitised at 20 kHz using an AD converter (PowerLab 26T, AD Instruments, USA) and acquired via commercial software (LabChart, AD Instruments, USA).

A custom-made non-invasive electrode was used to stimulate the right antenna. The electrode was placed 5 mm away from the base of the antenna. The flagellum of the antenna was inserted in a U-shaped cavity of the electrode and gently pressed to secure the physical contact between the antenna and the non-invasive electrode (Fig. 3a). The backpack was used to provide an electrical stimulation consisting of a single bipolar square-wave pulse at 1 Hz and a 50% duty cycle (Fig. 3b). The stimulus amplitude changed from 0.5 to 5.0 V with a 0.5 V step in an ascending manner. Each stimulation was repeated three times. A series of electrical stimulation was delivered to the antenna without conductive glue, and the same stimulation was repeated after dropping the conductive glue into the non-invasive electrode.

During the electrical stimulation of the antenna, the neural recording was contaminated by a stimulus artefact. To remove the effect of electrical stimulation, neural signals in a 40 ms period after the rising and falling edge of the square wave pulse were replaced with zero. Neural signals were then filtered with a second-order Butterworth filter (250 – 3,000 Hz). Detection of spike activities was performed by setting a threshold (Thr) as

*Thr = 5s*, with *s = median(|x|/0.6745)*        (1)

where x is the filtered signal [26]. Detected spikes, presumably multiunit activity, were used to calculate the mean spike count and coefficient of variation (CV). Because spike activities were elicited predominantly by the negative cycle of the stimulation (Fig. 3b), the number of spikes within 750 milliseconds after the falling edge of the pulse was counted as a neural response. Mean spike counts and CV was then calculated for individuals and compared between with-glue and without-glue condition. All analyses were carried out using MATLAB (Mathworks, USA).

*Fabrication of non-invasive electrodes and stimulation*

The non-invasive electrode used for the antenna was composed of electron conductors, Au-PI and an ion-conductive layer, PIL gel. Besides the PIL gel, a small 3D-printed fastener with a tunnel was designed to secure the contact between the electrode and the antenna.

The electron conductor was cut to strip. One end of the electron conductor strip was stuck to one side of the 3D-printed fastener with Super Gel from Yamayo Brand, and the other free end went through the connector's middle tunnel. A soft ring was then formed. The antenna was inserted into the ring until the end of the antenna. On the surface of the antenna end, the PIL gel with the optimal monomer ratio identified before was applied, and UV light was used for 100 seconds to make the gel more adhesive (Fig. 1c). The free end of the strip was pulled so that the ring could be tightened and firmly contact the surface of the antenna with the PIL gel. Super Gel was used to fix the stripe with a 3D-printed fastener at the end. A silver wire connected the electron conductor stripe and the pin. The pin was inserted into the backpack output channel (Fig. 1e).

The inter-space between the third and fourth segments was used for the grounding electrode. The inter-space between the sixth and seventh segments was used for the acceleration electrode. Both inter-spaces were filled with PIL gel, and the silver wire was inserted into the PIL gel (Fig. 1g). After the silver wire insertion, the tape was used to cover and fix the silver wire on the insect's third and sixth abdominal segments.

Fixed stimulation type and sequence were used to unify the stimulation on different insects. The turning and acceleration stimulations were set as 4V, 12ms bipolar wave and lasted 1 second, which was observed to elicit the insect's apparent related reaction.

*Experimental Setup for Locomotion Control*

A controller which can receive the workstation command and output the electrical stimulation was put on the back of the insect. The four implanted non-invasive electrodes were connected to the output channels of the controller. In order to verify whether the non-invasive electrodes can work for the turning and acceleration stimulation, three makers are put on the back of the insect so that the insect's movement can be recorded by the motion capture system and studied. The workstation saved the controller outputted electrical stimulation and the insect's motion data for the data analysis.

The cyborg insect was navigated along an "S" shaped path (Fig 5) to demonstrate the control superiority of the cyborg insect. The previous turning and acceleration stimulation parameters were applied for the manual navigation. The operator sent out the command of the intended stimulation type from the workstation to a central board which could transfer the command to the backpack of the cyborg insect via Bluetooth. The cyborg insect was then navigated to walk along the designed path.

24. Yu, Z. & Wu, P. Water-Resistant Ionogel Electrode with Tailorable Mechanical Properties for Aquatic Ambulatory Physiological Signal Monitoring. *Adv. Funct. Mater.* **31**, 2107226 (2021).

25. Chowański, S. *et al.* The physiological role of fat body and muscle tissues in response to cold stress in the tropical cockroach Gromphadorhina coquereliana. *PLOS ONE* **12**, e0173100 (2017).

26. Rey, H. G., Pedreira, C. & Quian Quiroga, R. Past, present and future of spike sorting techniques. *Brain Res. Bull.* **119**, 106–117 (2015).
**Acknowledgements**

This work was supported by the Singapore Ministry of Education (RG140/20). F.Z. and X.C. acknowledge financial support from the National Research Foundation Singapore (NRF) under NRF's Medium Sized Centre: Singapore Hybrid-Integrated Next-Generation µ-Electronics (SHINE) Centre funding programme. The authors appreciate Mr. Chong Bing Sheng for proofreading the manuscript, Mr. Roger Tan Kay Chia, and Ms. Kerh Geok Hong for their support.
**Author contributions**

Q.L., R.L., F.Z., X.C. and H.S. conceived and designed the research. Q.L., R.L. established insect control protocol. Q.L., R.L., K.K. and Z.C.O conducted the experiment and Q.L., K.K. did the analysis. F.Z. synthesized and characterized the non-invasive electrodes. Q.L., R.L., F.Z., K.K., X.C. and H.S. wrote and edited the manuscript. H.S. and X.C. supervised the research. All the authors read and revised the manuscript.

**Competing interests**

The authors declare no competing interests.

**Data availability**

The data that support the plots in this paper and other findings of this study are available from the corresponding authors upon reasonable request.

**Additional information**

Correspondence should be addressed to H.S. and X.C..

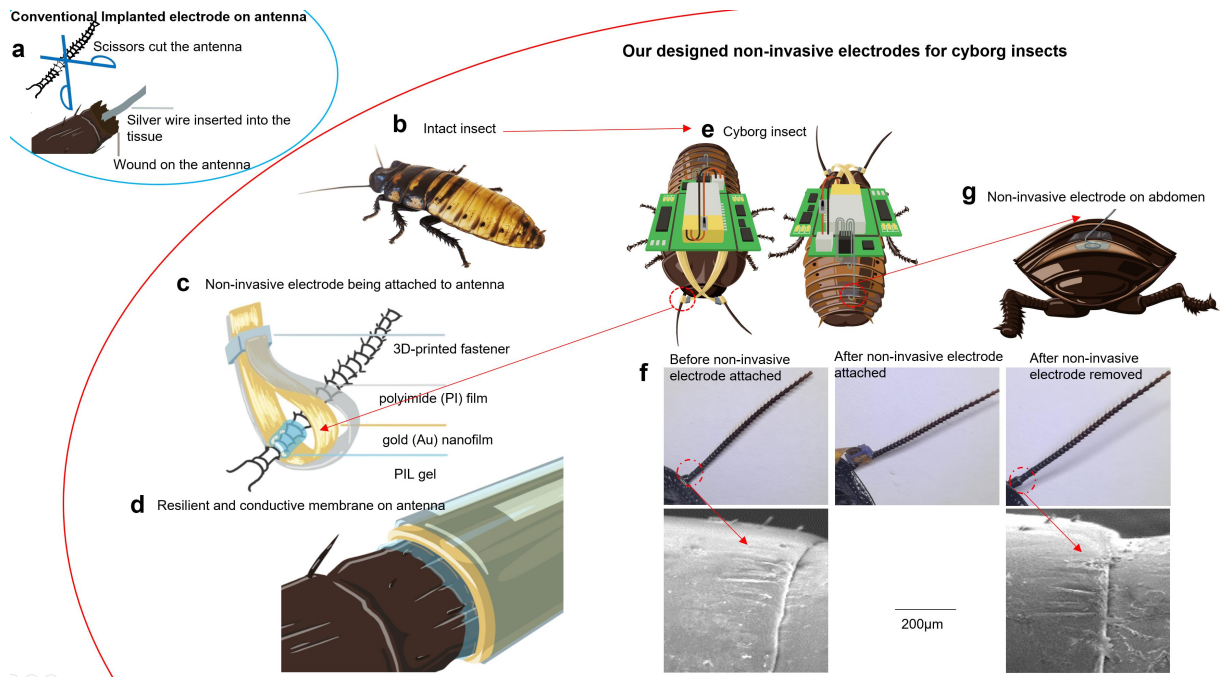

**Fig. 1 | Non-invasive electrodes with resilient and conductive membrane for cyborg insects. a,** The conventional way to implant an electrode on the antenna. The antenna should be cut before the conductor wire implantation as cutting and insertion lead to irreversible damage to the insect. The conductor silver wire was inserted on the antenna wound. **b,** An intact insect (Madagascar Hissing Cockroach) was used to build up the cyborg insect. **c,** Non-invasive electrode for the antenna. A 3D-printed fastener was used to tighten the contact between the antenna surface and the conductor (gold nanofilm and PIL gel) so that the conduct of electrical stimulation was stable and secured. Then, PIL gel was applied between the antennae and the gold nanofilm with 100 seconds of UV light for in-situ polymerisation. **d,** The insect's antenna surface was rough and uneven, while the proposed non-invasive electrode was conformable and flexible, which could come into contact firmly with the target surface of the antenna. In this case, the conduct of electrical stimulation could be secure and stable. **e,** The front and back view cyborg insect with non-invasive wearable electrodes attached to the antennae and abdomen. **f,** Antenna status before/after attaching and after removing the non-invasive electrodes. The antenna's intactness was preserved, indicating no

damage to the antenna. **g,** A non-invasive electrode is located on the insect's abdomen. For the non-invasive electrode on the insect's abdomen, PIL glue was firstly filled in the inter-space between two abdominal segments of the insect, and after a silver wire with a circular tip was inserted into the PIL gel, UV light was used for about 100 seconds for in-situ polymerisation. The silver wire's circular tip was to avoid injuries to insect skin. To secure the silver wire inside the PIL gel and to avoid the relative displacement of the silver wire from the insect's body, tape was used to fix the silver wire on the abdominal segment of the insect.

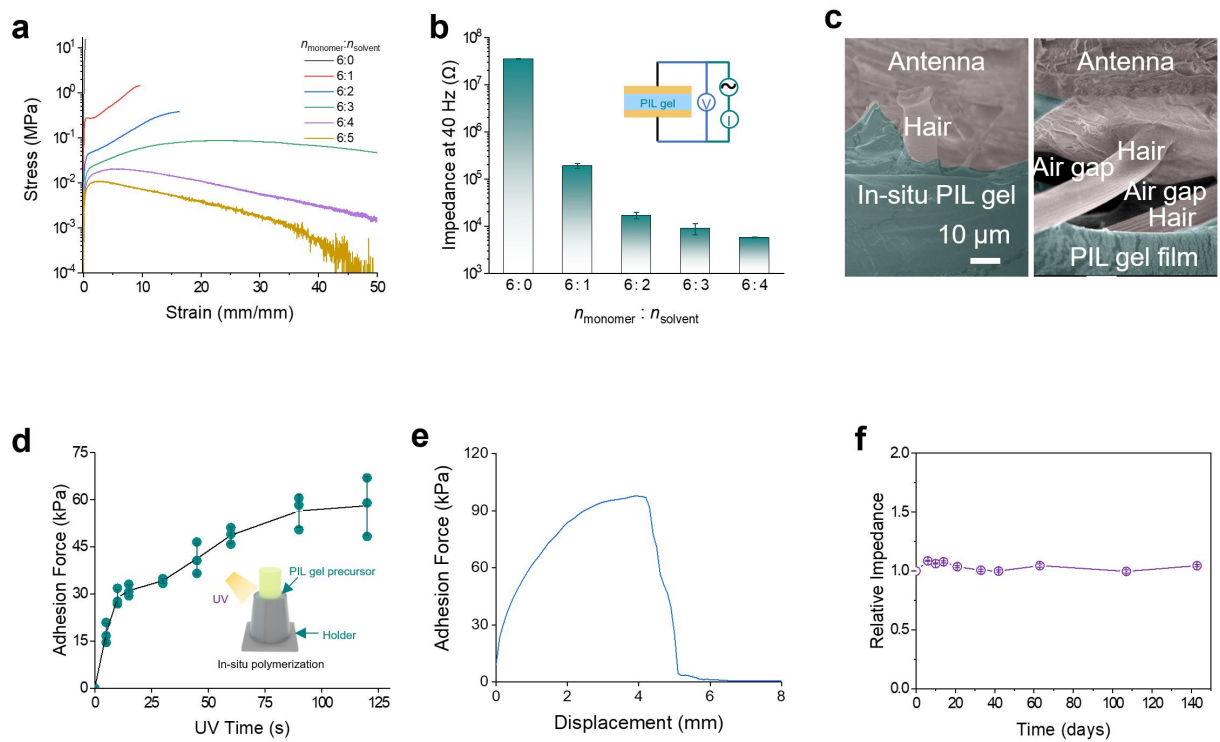

**Fig. 2 | Non-invasive electrodes of PIL gels by in-situ polymerisation. a,** Stress-strain curve of the PIL film with different solvent contents. The solvent increases the stretchability of the PIL gel but reduces its strength. **b,** The impedance of PIL gel with different monomer ratios at the frequency of 40 Hz, which is around the standard frequency of electrical stimulation on insects. The impedances of the PIL gels decrease as the solvent contents increase and reache a plateau at the $n_{monomer} : n_{solvent}$ of 6 : 2. **c,** The contact situation between PIL gel and antenna. In-situ polymerized PIL gel had a fuller contact with the antenna surface than the PIL gel film. **d,** The change of PIL gel adhesion force with UV time. Inset: a small setup simulated the antenna with PIL in-situ polymerisation to study the adhesion force. The adhesion force increases with UV exposure time and reaches a plateau at 100 s. **e,** The adhesion force between the antenna and the in-situ PIL gel. **f,** The long-term electrical stability ofthe PIL gels. The relative impedance almost unchanged during all these 140 days.

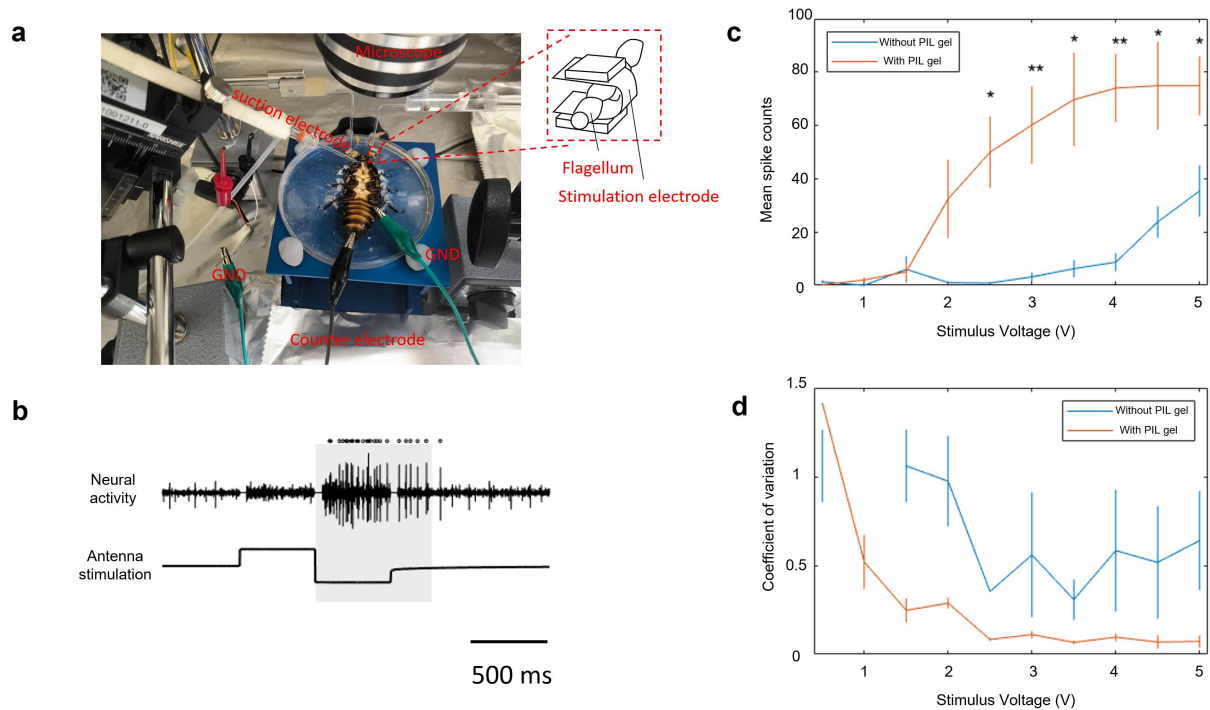

**Fig. 3 | Neural response to antenna stimulation with non-invasive electrodes. a,** Experiment setup. The insect was fixed at the platform. Its antennae were wrapped in the non-invasive electrodes. Neural recording on the neck nerve was conducted to check the insect's reaction to the electrical stimulation without or with PIL gel. **b,** Neural activity and antenna stimulation. Black dots denote detected neural spikes. In addition, neural reaction spikes in the shaded area were analysed. **c,** Mean of multiunit activities (N = 4). Asterisks represent statistical significance (*, $p < 0.05$. **, $p < 0.01$). The neural reaction spikes induced by the electrical stimulation with PIL gel were generally much more than the case without PIL gel, which shows the necessity of PIL gel for electrical stimulation. **d,** CV of multiunit activity (N = 3). CV is not applicable for 1 V stimulation without PIL gel since no spike unit was elicited. Generally, the insect's neural reaction spikes without PIL gel were not stable compared with the case with PIL gel. Meanwhile, with PIL gel, the insect's neural reaction spikes tend to stabilise after stimulus voltage increases over 2.5 v.

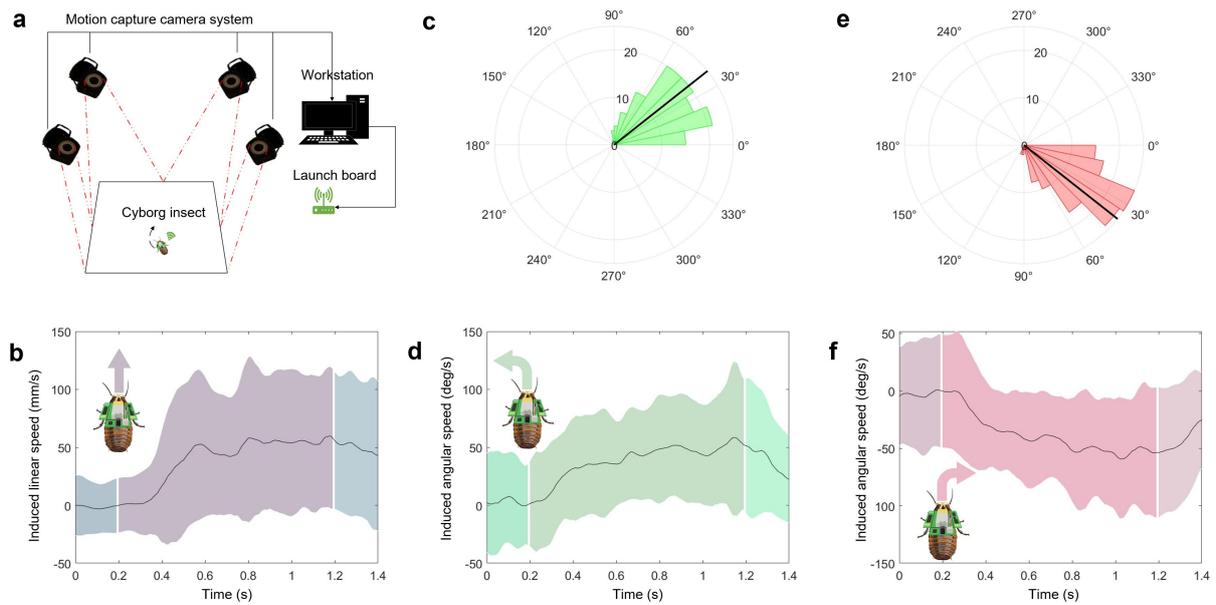

**Fig. 4 | Insect's motion induced by the electrical stimulation with non-invasive electrodes. a,** Motion Tracking System for the cyborg insect's locomotion control. The Vicon camera system feedbacked the insect's position to the workstation while the workstation send out the stimulation type to the cyborg insect through a launchboard. **b,** Acceleration stimulus. The induced linear speed increased from 0 mm/s to 60 mm/s during the stimulation and decreased after the stimulation stopped. **c,** Left-turning angle. The distribution of the successfully induced left turning angle. Based on the success criteria, the success rate of the left tuning stimulation is 80.7%. The induced turning angles varied from 0 to around 105.5 degrees, while most were between 0 and 60 degrees. Averagely, the left turning angle was 38.5 degrees. **d,** Left-turning stimulus. The induced Left turning angular speed fluctuated from 0 to around 60 degrees/s and decreased after the stimulation stopped. **e,** Right-turning angle. The distribution of the successfully induced right turning angle. Based on the criteria for successful right-turning stimulation, the success rate of the right-turning stimulation is 80.7%. The induced turning angles varied from 0 to around 108.6 degrees, while the majority were between 0 and 60 degrees. Averagely, the left-turning angle was 38.4 degrees. **f,** Right-turning stimulus. The induced right-turning angular speed increased from 0

to around 60 degree/s before the stimulation stopped and then fluctuated until the stimulation stopped. Afterwards, the induced angular speed started to decrease.

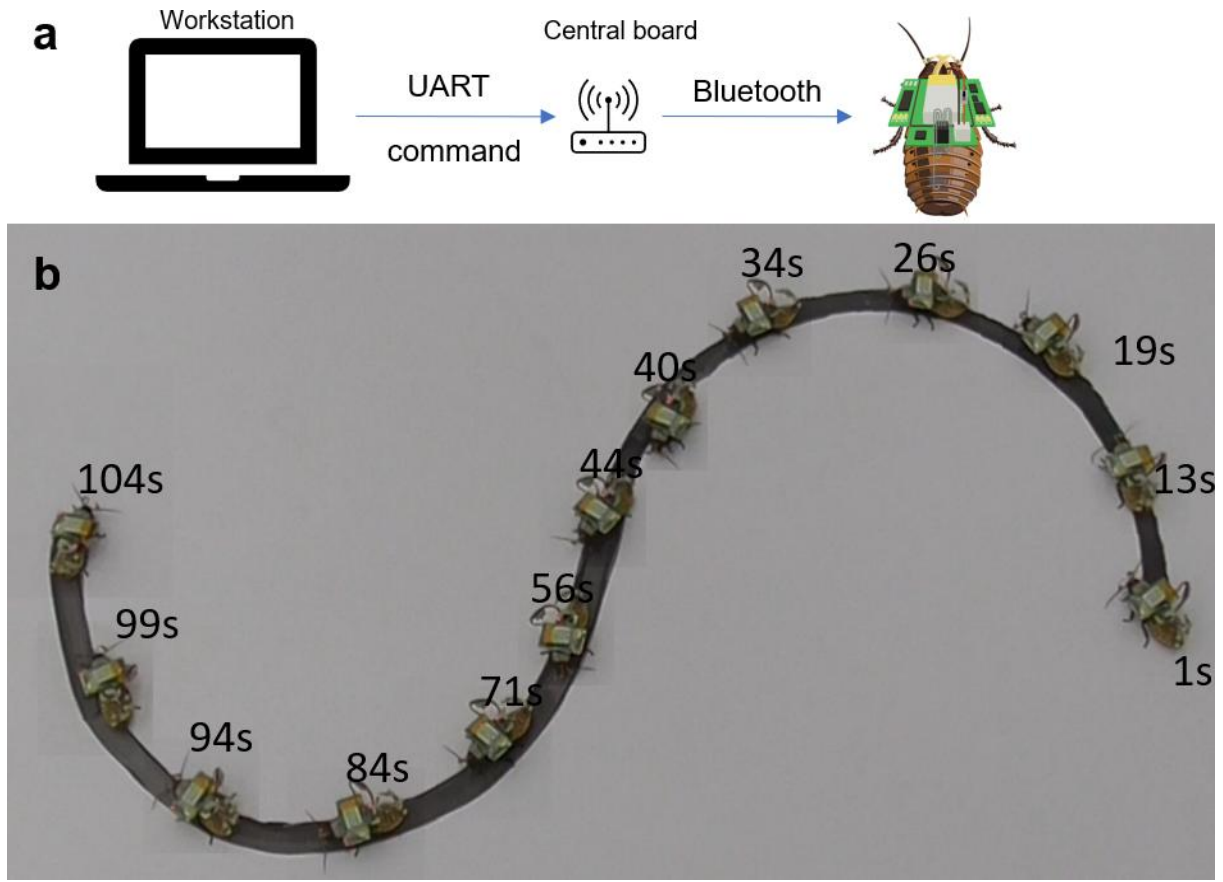

**Fig. 5 | "S" shape path following behaviour exhibited by the cyborg insect. a,** The control procedure of cyborg insects. The operator sent out the command of stimulation type from the workstation. The command was then sent to the central board via UART and transferred to the cyborg insect via Bluetooth. The cyborg insect then reacted based on the command. **b,** "S" path following demonstration of the cyborg insect. The insect was manually navigated along the "S" path with the stimulation parameter verified before. As a result, its movement trajectory was close to the designed path, indicating its advantages in controllability.